%% file: main.tex
\documentclass[10pt,conference,a4paper]{IEEEtran}

\usepackage{cite}
\usepackage{graphicx}
\usepackage{amsmath,amssymb,amsfonts}
\usepackage{algorithmic}
\usepackage{array}
\usepackage{url}
\usepackage{textcomp}
\usepackage{xcolor}
\usepackage{comment}
\usepackage{tabu}
\usepackage{booktabs}
\usepackage{multirow}
\usepackage{lipsum}
\usepackage[bookmarks=false]{hyperref}
\usepackage{amssymb}
\usepackage{pifont}
\usepackage{multicol}
\usepackage{pbox}
\usepackage{float}

\input{text/0_macro}

\def\BibTeX{{\rm B\kern-.05em{\sc i\kern-.025em b}\kern-.08em
    T\kern-.1667em\lower.7ex\hbox{E}\kern-.125emX}}

\begin{document}
\title{The color out of space: learning self-supervised representations for Earth Observation imagery}
\author{
\IEEEauthorblockN{
Stefano Vincenzi\IEEEauthorrefmark{1}, 
Angelo Porrello\IEEEauthorrefmark{1}, \
Pietro Buzzega\IEEEauthorrefmark{1}, \
Marco Cipriano\IEEEauthorrefmark{1},
Pietro Fronte\IEEEauthorrefmark{3}, \\
Roberto Cuccu\IEEEauthorrefmark{3}, \
Carla Ippoliti\IEEEauthorrefmark{2},\  
Annamaria Conte\IEEEauthorrefmark{2},\  
Simone Calderara\IEEEauthorrefmark{1}}
\\
\IEEEauthorblockA{ 
\IEEEauthorrefmark{1}AImageLab, University of Modena and Reggio Emilia, Modena, Italy\\
\IEEEauthorrefmark{2}Istituto Zooprofilattico Sperimentale dell'Abruzzo e del Molise `G.Caporale', Teramo, Italy}
\IEEEauthorrefmark{3}Progressive Systems Srl, Frascati -- Rome, Italy\\
}
\maketitle
\begin{abstract}
\input{text/0_abstract}
\end{abstract}
\IEEEpeerreviewmaketitle
\input{text/1_introduction}
\input{text/2_related}
\input{text/3_model}
\input{text/4_datasets}
\input{text/5_experiments}
\input{text/6_conclusions}
\section*{Acknowledgment}
The research described in this paper has been conducted within the project `AIDEO' (AI and EO as Innovative Methods for Monitoring West Nile Virus Spread). The project is being developed within the scope of the ESA EO Science for Society Permanently Open Call for Proposals EOEP-5 BLOCK 4 (ESA AO/1-9101/17/I-NB). 
%
%
\bibliographystyle{IEEEtran}
\bibliography{main}

\end{document}

%% file: text/0_macro.tex
\newcommand\mytabheader[1]{\bf{#1}}

\newcommand{\resnet}[1]{ResNet-#1}

\newcommand{\nicepar}[1]{\noindent \normalfont\textbf{#1}}

\newcommand{\quotationmarks}[1]{``#1''}

\newcommand{\tinyrgbGT}[1]{\includegraphics[width=0.21\columnwidth]{imgs/colors_grad/rgbGT_#1.png}}
\newcommand{\tinypred}[1]{\includegraphics[width=0.21\columnwidth]{imgs/colors_grad/prediction_#1.png}}
\newcommand{\tinyrgbheat}[1]{\includegraphics[width=0.21\columnwidth]{imgs/rgb_heatmaps/heat_#1.png}}
\newcommand{\tinyspectralheat}[1]{\includegraphics[width=0.21\columnwidth]{imgs/spectral_heatmaps/heat_#1.png}}

\newcommand{\bigearth}{BigEarthNet}

%% file: text/0_abstract.tex
The recent growth in the number of satellite images fosters the development of effective deep-learning techniques for Remote Sensing (RS). However, their full potential is untapped due to the lack of large annotated datasets. Such a problem is usually countered by fine-tuning a feature extractor that is previously trained on the ImageNet dataset. Unfortunately, the domain of natural images differs from the RS one, which hinders the final performance. In this work, we propose to learn meaningful representations from satellite imagery, leveraging its high-dimensionality spectral bands to reconstruct the visible colors. We conduct experiments on land cover classification (BigEarthNet) and  West Nile Virus detection, showing that colorization is a solid pretext task for training a feature extractor. Furthermore, we qualitatively observe that guesses based on natural images and colorization rely on different parts of the input. This paves the way to an ensemble model that eventually outperforms both the above-mentioned techniques.

%% file: text/1_introduction.tex
\section{Introduction}
\label{sec:Intro}

Over the last decades, Remote Sensing has become an enabling factor for a broad spectrum of applications such as disaster prevention~\cite{schumann2018assisting}, wildfire detection~\cite{filipponi2019exploitation}, vector-borne disease~\cite{ippoliti2019defining}, and climate change~\cite{rolnick2019tackling}. These applications benefit from a higher number of satellite imagery captured at unprecedented rhythms~\cite{drusch2012sentinel}, thus making every aspect of the Earth's surface constantly monitored. Machine learning and Computer Vision provide valid tools to exploit these data in an efficient way. Indeed, a synergy between Earth Observation and Deep Learning techniques led to promising results, as highlighted by recent advances in land use and land cover classification~\cite{ji20183d}, image fusion~\cite{fusion_1}, and semantic segmentation~\cite{segmentation_1}. 

Despite the amount of raw information being significant, the exploitation of these data still raises an open problem. Indeed, the prevailing learning paradigm -- the supervised one -- frames the presence of labeled data as a crucial factor. However, acquiring a huge amount of ground truth data is expensive and requires expert staff, equipment, and in-field measurements. This often restrains the development of many downstream tasks that are important for paving the way to the above-mentioned applications.

To mitigate such a problem, a common solution~\cite{huh2016makes} exploits models that are pre-trained on the ImageNet~\cite{imagenet} dataset. In detail, the learning phase is conducted as follows: firstly, a deep network is trained on ImageNet until it reaches good performance on image categorization; secondly, a fine-tuning step is carried out on a target task (\textit{e.g.} land cover classification). This way, one can achieve acceptable results even in the presence of few labeled examples, as the second step just adapts a set of general-purpose features to the new task. However, this approach is limited only to the tasks involving RGB images as input. Satellite imagery represents a domain that is quite different from the RGB one, thus making the ImageNet pre-training only partially suitable.

These considerations reveal the need for novel approaches that are tailored for satellite imagery. To build transferable representations, two kinds of approaches arise from the literature: annotation-based methods and self-supervised ones. The authors of~\cite{in_domain} fulfill the principle of the first branch by investigating in-domain representation learning. They shift the pre-training stage from ImageNet to a labeled dataset specific for remote sensing. As an example, one could leverage BigEarthNet~\cite{sumbul2019bigearthnet}, which has been recently released for land-cover classification. On the other hand, Tile2Vec~\cite{jean2019tile2vec} extracts informative features in a self-supervised fashion. The authors rely on the assumption that spatially close tiles share similar information: therefore, their corresponding representations should be placed closer than tiles that are far apart. In doing so, one does not need labeled data for extracting representations, but lacks robustness when close tiles are not similar.

Similarly to~\cite{jean2019tile2vec}, we propose a novel representation learning procedure for satellite imagery, which devises a self-supervised algorithm. In more detail, we require the network to recover the RGB information by means of other spectral bands solely. For the rest of the article, we adopt the term \quotationmarks{spectral bands} for indicating the subset of the bands not including the RGB. Our approach closely relates to colorization, which turns out to encourage robust and high-level feature representations~\cite{larsson2017colorization,zhang2017split}. We feel this pretext task being particularly useful for satellite imagery, as the connection between colors and semantics appears strong: for instance, sea waters feature the blue color, vegetation regions the green one or arable lands prefer warm tones. We inject such a prior knowledge through an encoder-decoder architecture that -- differently from concurrent works -- exploits spectral bands (\textit{e.g.} short-wave infrared, near-infrared, etc.) instead of grayscale information to infer color channels. Once the model has reached good capabilities on tile colorization, we use its encoder as a feature extractor for the later step, namely fine-tuning on a remote sensing task. We found that the representations learnt by colorization leads to remarkable results and semantically diverge from the ones computed on top of RGB channels. Taking advantage of these findings, we set up an ensemble model, which averages the predictions from two distinct branches at inference time (the one fed with spectral bands, the other with RGB information). We show that ensembling features this way leads to better results. To the best of our knowledge, our work is the first investigating colorization as a guide towards suitable features for remote sensing applications.

To show the effectiveness of our proposal, we assess it in two different settings. Firstly, we conduct experiments on land-cover classification, comparing our solution with two baselines, namely training from scratch and fine-tuning the ImageNet pre-training. We show that colorization is particularly effective when few annotations are available for the target tasks. This makes our proposal viable for scenarios where gathering many labeled data is not practicable. To demonstrate such a claim, we additionally conduct experiments on the \quotationmarks{West Nile Virus} cases collected in the frame of the Surveillance plan put in place by the Ministry of Health, with the aim of predicting presence/absence across the Italian territory.

%% file: text/2_related.tex
\section{Related Works} \label{sec:related}
\subsection{Land cover - Land use classification}

Recently, the categorization of land-covers has attracted wide interest, as it allows for the collection of statistics, activities planning, and climate changes monitoring. To address these challenges, the authors of~\cite{related_cnn_1} exploit Convolutional Neural Networks (CNN) to extract representations encoding both spectral and spatial information. To speed up the learning process, they advocate for a prior dimensionality reduction step across the spectra, as they observe a high correlation in this dimension. Among works focusing on how to exploit spectral bands,~\cite{related_rnn} devises Recurrent Neural Networks (RNNs) to handle the redundancy underlying adjacent spectral channels. Similarly,~\cite{related_cnn_4} proposes a 3D-CNN framework, which can naturally joint spatial and spectral information in an end-to-end fashion without requiring any pre-processing step.

While these approaches concern the design of the feature extractor, our work is primarily engaged in the scenarios in which few labeled examples are available. In these contexts, fine-tuning pre-trained models often mitigate the lack of a large annotated dataset, yielding great performance in some cases~\cite{tuning_1,tuning_2}. Intuitively, the representations learned from ImageNet ($1$ million images belonging to $1000$ classes) encode a prior knowledge on natural images, thus facilitating the transfer to different domains. Instead,~\cite{in_domain} proposes in-domain fine-tuning, where the pre-training stage performs on a remote sensing dataset. The authors found in-domain representations to be especially effective with limited data ($1000$ training examples), surpassing the performance yielded by the ImageNet initialization. As a final remark, one could reduce overfitting through data augmentation~\cite{aug_1} (\textit{i.e.} flip, translation, and rotation), which increases both the diversity and volume of training data.
 
\subsection{Unsupervised Representations Learning}

Unsupervised and self-supervised methods were introduced to learn general visual features from unlabeled data~\cite{jing2019self}. These approaches often rely on \textit{pretext tasks}, which attempt to compensate for the lack of labels through an artificial supervision signal. In so doing, the learned representations hopefully embody meaningful information that is beneficial to downstream tasks.

\nicepar{Reconstructions-based methods}. Under this perspective, generative models can be considered as self-supervised methods, where the reconstruction of the input acts as a pretext task. Denoising Autoencoders~\cite{vincent2008extracting} contribute to this line of research: here, the learner has to recover the original input from a corrupted version. The idea is that good representations are those capturing stable patterns, which should be recovered even in the presence of a partial or noisy observation. In remote sensing, autoencoders are often applied~\cite{AE_1,AE_2,related_cnn_1} to reduce the dimensionality of the feature space. This yields the twofold advantage of decreasing the correlation lying in spectral bands and reducing the overall computational effort.

\nicepar{Classification-based methods}.~\cite{gidaris2018unsupervised} frames the pretext task as a classification problem, where the learner guesses which rotation ($0$°, $90$°, $180$° and $270$°) has been applied to its input. The authors observe that recognizing the input transformation behaves as a proxy for object recognition: the higher the accuracy on the upstream task, the higher the accuracy on the downstream one. Considering two random patches from a given image,~\cite{doersch2015unsupervised} asks the network to infer the relative position between those. This encourages the learner to recognize the parts that make up the object as well as their relations. Similarly,~\cite{noroozi2016unsupervised} presents a jigsaw puzzle to the network, which has to place the shuffled patches back to their original locations.

\nicepar{Colorization-based methods}. Given a grey-scale image as input, colorization is the process of predicting realistic colors as output. A qualitative analysis conducted in~\cite{col_7} shows that colorization-driven representations capture semantic information, grouping together high-level objects that display low-level variations (\textit{e.g.} color or pose).~\cite{col_5} concerns the ambiguity and ill-posedness of colorization, arguing that several solutions may be assessed for a given grey-scale image. On this basis, the authors exploit Conditional Variational Autoencoder (CVAE) to produce diverse colorizations, thus naturally complying with the multi-modal nature of the problem. Instead,~\cite{col_4} focuses on the design of the inference pipeline and proposes a two-stage procedure: \textit{i)} a pixel-wise descriptor is built by VGG-16 feature maps taken at different resolutions; \textit{ii)} the descriptors are then fed into a fully connected layer, which outputs hue and chroma distributions. 
Split-Brain Autoencoders~\cite{zhang2017split} relies on a network composed of two disjoint modules, each of which predicts a subset of color channels from another. The authors argue that this schema induces transferable representations, the latter taking into account all input dimensions (instead of gray-scale solely).

%% file: text/3_model.tex
\section{Model}
\begin{figure*}[t]
    \centering
    \includegraphics[width=\textwidth]{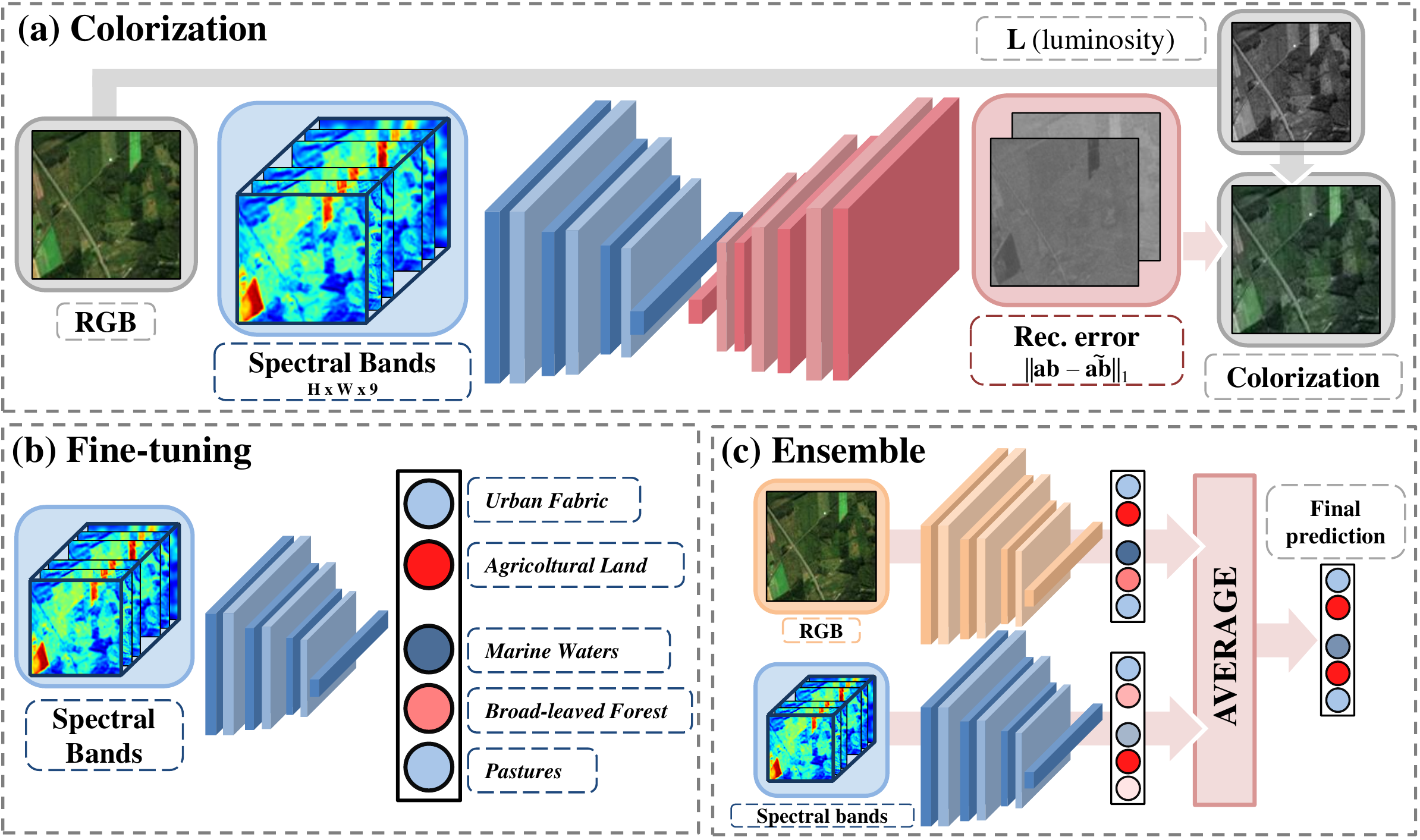}
    \caption{An overview of the proposed pipeline for feature learning on satellite imagery.}
    \label{fig:framework}
\end{figure*}
\nicepar{Overview}. Our main goal consists in finding a good initialization for the classifier, in such a way that it can later capture meaningful and robust patters even in presence of few labeled data. To this purpose, we devise a two-stage procedure tailored for satellite imagery tasks, which prepends a colorization step (Sec.~\ref{sec:colorization}) to a fine-tuning one (Sec.~\ref{sec:finetuning}).

As depicted in Fig.~\ref{fig:framework}~(a), our proposal leverages an encoder-decoder architecture for feature learning. In doing so, we do not require the model to reconstruct its input: differently, we set up an asymmetry between input (spectral bands) and output (color channels). This way, we expect the encoder to capture meaningful information about soil and environmental characteristics. Afterward, we exploit the encoder and its representation capabilities to tackle a downstream task (\textit{e.g.} land cover classification, see Fig.~\ref{fig:framework}~(b)). Eventually, an ensemble model (see Sec.~\ref{sec:ensemble} for additional details) further refines the final prediction combining the outputs from the two input modalities (RGB and spectral bands). 

\subsection{Colorization} 
\label{sec:colorization}

In formal terms, the encoder network $\mathcal{F}$ takes $\mathbf{S} \in \mathbb{R}^{H\times W \times C}$ as input, where $C$ equals the number of spectral bands available to the model and $H$ and $W$ the input resolution (height and width respectively). The decoder network produces a tensor $\widehat{\mathbf{X}}\in\mathbb{R}^{H\times W \times 2}$, which yields the pixel-wise predictions in terms of $a$ and $b$ coordinates in the CIE \textit{Lab} color space. On this latter point, a naive strategy would simply define the expected output in terms of RGB: nevertheless, as pointed out in~\cite{col_4}, modeling colors as RGB values may not yield an effective training signal. Differently, we adhere to the guideline described in~\cite{zhang2016colorful} and frame the problem in the CIE \textit{Lab} space. Here, a color is defined with a lightness component $L$ and $a*b$ values carrying the chromatic content. The effectiveness of this space comes from the fact that colors are encoded accordingly to human perception: namely, the distance between two points reflects the amount of visually perceived change between the corresponding colors.

\nicepar{Encoder}. We opt for ResNet18~\cite{he2016deep} as backbone network for the encoder, which hence consists of four blocks with two residual units each. As pointed out in~\cite{kolesnikov2019revisiting}, thanks to their residual units and skip connections, ResNet-based networks are more suitable for self-supervised representation learning. Indeed, when compared to other popular architectures (\textit{e.g.} AlexNet), residual networks favorably preserves representations from degrading towards the end of the network and therefore results in better performance.

\nicepar{Decoder}. In designing the decoder network, we mirror the architecture of the encoder, replacing the first convolutional layer of each residual block with its transposed counterpart. Moreover, we add an upsampling operation to the top of the decoder, followed by a batch normalization layer, a ReLU activation, and a transposed convolution. The latter reduces the number of features maps to $2$: this way, the output dimensionality matches the ground truth one.

\nicepar{Colorization Loss}. Recent works~\cite{zhang2017split, zhang2016colorful, col_7} investigate various loss functions, questioning their contributions to colorization results (intended as performance on either the target task or the pretext one). Despite a regression objective (\textit{e.g.} the mean squared error) being a valid baseline, these works show that treating the problem as a multinomial classification leads to better results. However, the overall training time increases considerably because of the additional information taken into account. In our case, this would add up to the burdensome computations required by hyperspectral images, thus resulting even more expensive. For this reason, we limit our experiments to the mean absolute error $\mathcal{L}_{1}(\cdot, \cdot)$, as follows:
\begin{equation}
\mathcal{L}_{1}(\mathcal{\widehat{\mathbf{X}}}, \mathbf{X}) = \lambda \sum_{h,w} \left|{\widehat{x}_{h,w}^{\ (a)}-x_{h,w}^{\ (a)}}\right| + \left|{\widehat{x}_{h,w}^{\ (b)}-x_{h,w}^{\ (b)}}\right|,
\label{eq:l1loss}
\end{equation}
where $\mathbf{X}$ represents the $a*b$ ground truth colorization and $\lambda = 100$ is a weighting term that prevents numerical instabilities.

\subsection{Fine-tuning}
\label{sec:finetuning}

Once the encoder-decoder has been trained, we turn our attention to the downstream task and exploit the encoder $\mathcal{F}(\cdot)$ as a pre-trained feature extractor. To achieve this, we need a single amendment to the network: a final linear transformation that maps bottleneck features $\mathbf{H} = \mathcal{F}(\mathbf{S})$ to the classification output space $\widehat{\mathbf{y}} = \mathbf{W}^T \mathbf{H} + \mathbf{b}$.

\nicepar{Classification Loss} We make use of two different losses in our experiments: when dealing with a multi-label task as the land cover classification one (\textit{i.e.} each example can be categorized into multiple classes), the objective function resembles a binary cross-entropy term averaged over C classes:
\begin{equation}
    \mathcal{L}(\mathcal{\widehat{\mathbf{y}}}, \mathbf{y}) = -\frac{1}{C} \sum_{i} \mathbf{y}_{i} \log \sigma \left( \widehat{\mathbf{y}}_{i}\right) + (1-\mathbf{y}_{i}) \log \left( 1 - \sigma \left(\widehat{\mathbf{y}}_{i}\right) \right),
    \nonumber
\end{equation}
where $\mathbf{y}$ indicates the ground-truth multi-hot encoding vector and $\sigma$ the sigmoid function. Differently, we use the binary cross-entropy loss to treat the West Nile Disease case study. 

\subsection{Model ensemble}
\label{sec:ensemble}

As pointed out in~\cite{zhang2017split}, a network trained on colorization specializes just on a subset of the available data (in our case, spectral bands) and cannot exploit the information coming from its ground truth (the RGB color images). To further take advantage of color information, we set up an ensemble model at inference time (so, no additional training steps required). As shown in Fig.~\ref{fig:framework}~(c), the ensemble is formed by two independent branches taking the RGB channels and the spectral bands as input respectively. The first one is pre-trained on classification (ImageNet) and the second one on colorization; both are fine-tuned separately on the given classification task. The ensemble-level predictions are simply computed by averaging the responses from the two branches: 
\begin{equation}
    \widehat{\mathbf{y}}_{\text{ENS}} = \frac{\sigma(\widehat{\mathbf{y}}_{\text{RGB}}) + \sigma(\widehat{\mathbf{y}}_{\text{SPECTRAL}})}{2}.
\end{equation}

%% file: text/4_datasets.tex
\section{Datasets}
\label{sec:dataset}

The two datasets we rely on data acquired through the Sentinel-2A and 2B satellites developed by the European Space Agency (ESA). These satellites provide a multi-spectral imagery over the earth with 12 spectral bands (covering the visible, near and short wave infrared part of the electromagnetic spectrum) at three different spatial resolutions ($10$, $20$ and $60$ meters per pixel).

\subsection{Land-cover classification - BigEarthNet}

In Remote Sensing, the main bottleneck in the adoption of deep networks was the lack of a large training set. Indeed, existing datasets (as Eurosat~\cite{helber2019eurosat}, PatterNet~\cite{zhou2018patternnet}, UC Merced Land Use Dataset~\cite{yang2010bag}) include a small number of annotated images, hence resulting inadequate for training very deep networks. To overcome this problem,~\cite{sumbul2019bigearthnet} introduces BigEarthNet, a novel large scale dataset collecting $590\,326$ tiles. Each example comprises of $12$ bands (RGB included) and multiple land-cover classes (provided by the CORINE Land Cover (CLC) database~\cite{feranec2016european}) as ground truth. 

Originally, the number of classes amounted to $43$: but, the authors of~\cite{sumbul2020bigearthnet} argue that some CORINE classes cannot be easily inferred by looking at Sentinel-2 images solely. Indeed, some labels may not be recognizable at such low resolution (the highest one is $120\times120$ pixels for $10m$ bands) and other ones would require temporal information for being correctly discriminated (\textit{e.g.} non-irrigated arable land \textit{vs.} permanently irrigated land). For these reasons, in our experiments we adopt the class-nomenclature proposed in~\cite{sumbul2020bigearthnet}, which reduces the number of classes to $19$. Moreover, we discard the $70\,987$ patches displaying lands that are fully covered by clouds, cloud shadows, and seasonal snow.

\subsection{West Nile Disease Dataset}
\label{sec:westnile}

In the last decade, numerous studies have examined the complex interactions among vectors, hosts, and pathogens~\cite{ippoliti2019defining,tran2014environmental}. In particular, one of the major threat worldwide studied is represented by West Nile Disease (WND), a mosquito-borne disease caused by West Nile virus (WNV). Mosquitoes presence and abundance have been extensively proved to be associated with climatic and environmental factors such as temperatures, vegetation, rainfall~\cite{tran2014environmental,bisanzio2011spatio,conte2015spatio}, and remote sensing has been an important key source for data collection. Our capacity to collect and store data continues to expand rapidly and this requires the incorporation of new analytical techniques able to process Earth Observation (EO) data establishing pipelines to turn near real-time “big data” into “smart data”~\cite{vincenzi2019spotting}. In this context, Deep techniques could provide useful tools to process data and automatically identify patterns able to make accurate predictions of the spatio-temporal re-emergence and spread of the West Nile Disease in Italy. With this aim, we collected data from the Copernicus program and paired Sentinel 2 (S2) EO data with ground truth WND data. 

Disease sites are collected through the National Disease Notification System of the Ministry of Health (SIMAN www.vetinfo.sanita.it)~\cite{colangeli2011sistema}. We start with the analysis of the 2018 epidemic, one of the most spread on the Italian territory. We frame the problem as a binary classification task with the final purpose of predicting positive and negative WND sites analyzing multi-spectral bands. Positive cases are geographically located mainly in Po valley, in Sardinia and some spots in the rest of Italy~\cite{riccardo2018early}: the location of each case of birds, mosquitoes and horses, was visually inspected for the accuracy needs in the analysis. Negative sites, being not always available in the national database due to the surveillance plan strategy, were derived as pseudo-absence ground truth data, either in the space (points located in areas where the disease was never reported in the past) and in the time (a random date in months previous the reported positivity in mosquitoes collections).

WND dataset comprises of $1\,488$ distinct cases, divided into $962$ negatives and $526$ positives. Each case comes with a variable number of Sentinel-2 patches (corresponding to various acquisitions over time), thus leading to $18\,684$ spectral images in total.

%% file: text/5_experiments.tex
\section{Experiments}
\noindent In this section, we test our proposal as a pre-training strategy for the later fine-tuning step. We compare the results yielded by colorization to those achieved by two baselines: training from scratch~\cite{he2015delving} and the common ImageNet pre-training. In doing so, we mimic scenarios with few labeled data by reducing the amount of examples available at training time (\textit{e.g.} $1\,000$, $5\,000$, etc\dots).

\subsection{Evaluation Protocols}

\nicepar{Land-Cover Classification}. We strictly follow the guidelines provided by~\cite{in_domain} when assessing the performance on the \bigearth{} benchmark. Namely, we form the training set by sampling $60$\% of the total examples, retaining $20$\% for the validation set and $20$\% for the test set. We measure the results in terms of Mean-Average-Precision (mAP), which also considers the order in which predictions are given to the user. We check the performance every $10$ epochs and retain the weights that yield the higher mAP score on the validation set. 

\nicepar{West Nile Disease}. Here, we adopt the stratified holdout strategy, which ensures the class probabilities of training and test being close to each other. The metrics of interest are precision, recall and F1 score, the latter accounting for the slight imbalance that occurs at class level (indeed, negatives cases appear more frequently than positives ones).

\subsection{Implementation details}

\nicepar{\bigearth{}}. We exploit the normalization technique described in~\cite{prathap2018deep,vincenzi2019spotting} computing the $2$\textsuperscript{nd} and $98$\textsuperscript{th} percentile values to normalize each band. This method is more robust than the common min-max normalization, as it is less sensitive to outliers. Before feeding the spectral bands into the model -- as they come with different spatial resolutions -- we apply a cubic interpolation to get a dimension of $128 \times 128$.

\nicepar{Colorization}. To broaden the diversity of available data, we apply data augmentation (\textit{i.e.} rotation, horizontal and, vertical flip). We initialize the network according to~\cite{he2015delving} and train for $50$ epochs on the full BigEarthNet, setting the batch size equal to $16$ and leveraging Stochastic Gradient Descent (SGD) as optimizer (with a learning rate fixed at $0.01$).

\nicepar{Land-Cover Classification}. We train the model for $30$ epochs whether the full dataset is available; otherwise we increase the epochs to $50$. The learning rate is set to $0.1$ and divided by $10$ at the $10^{\text{th}}$ and $40^{\text{th}}$ epoch. The batch size equals $64$. 

\nicepar{West Nile Disease} Differently from the previous cases, we apply neither upscaling nor pixel-normalization, as all channels are provided at the same resolution ($224\times224$) and their values lie within the range $[0,1]$. We leverage the network trained for colorization on \bigearth{}. Since we rely on a subset of the spectral bands ($B_{1}$, $B_{8A}$, $B_{11}$ and $B_{12}$), we fix the first convolutional layer so that it takes $4$ channels as input. We optimize the model for $30$ epochs, with a batch size of $32$ and an initial learning rate of $0.001$, multiplied by $0.1$ after $25$ epochs. 

\subsection{Results of Colorization pre-training}

\begin{table}[t]
\caption{Performance (mAP) on \bigearth{} for different strategies to vary the number of training examples.}
    \begin{center}
        \begin{tabular}{lcccccc}
        \toprule
         \mytabheader{Input} & \mytabheader{pre-training} & \mytabheader{1k} & \mytabheader{5k} & \mytabheader{10k} & \mytabheader{50k} & \mytabheader{Full} \vspace{0.1cm}\\
         \midrule
         RGB & from scratch & .486 & .608 & .645 & .744 & .851 \\
         RGB & ImageNet & .620 & .695 & .726 & .786 & \textbf{.879} \\
         \midrule
         Spectral & from scratch  & .555 & .667 & .711 & .767 & .866 \\
         Spectral & ImageNet & .578 & .627 & .681 & .773 & \textbf{.879} \\
         Spectral & Color. (our) & \textbf{.622} & \textbf{.730} & \textbf{.760} & \textbf{.793} & .860 \\
         \bottomrule
        \end{tabular}
    \label{tab:bigresult}
    \end{center}
\end{table}
\begin{table}[t]
\caption{Ensemble model -- results (mAP) on \bigearth{}.}
    \begin{center}
        \begin{tabular}{lcccccc}
        \toprule
         \mytabheader{Input} & \mytabheader{pre-training} & \mytabheader{1k} & \mytabheader{5k} & \mytabheader{10k} & \mytabheader{50k} & \mytabheader{Full} \vspace{0.1cm}\\
         \midrule
         RGB & ImageNet & .620 & .695 & .726 & .786 & .879 \\
         \midrule
         Spectral & Colorization  & .622 & .730 & .760 & .793 & .860 \\
         \midrule
         Ensemble & ImagNet+ImageNet & .649 & .707 & .749 & .815 & \textbf{.904} \\
         Ensemble & Color.+ImageNet & \textbf{.656} & \textbf{.751} & \textbf{.778} & \textbf{.823}  & .896 \\
         \bottomrule
        \end{tabular}
    \label{tab:ensembleresult}
    \end{center}
\end{table}

Based on the final performance reported in Tab.~\ref{tab:bigresult}, one can observe the initialization offered by colorization surpassing the other alternatives. Such a claim especially holds in presence of scarce data, thus complying with the goals we have striven for in this work. This does not apply when the learner faces up to the entire training set ($519$k examples): such evidence -- already encountered in~\cite{in_domain} -- deserves more investigations that we will conduct in future works.

Results shown by Tab.~\ref{tab:bigresult} let us draw additional remarks: \textit{i)} as one would expect, the ImageNet pre-training performs good for an RGB input; however, when dealing with the spectral domain, even a random initialization outperforms it; \textit{ii)} colorization is the sole that rewards the exploitation of spectral bands and justifies their usage in place of RGB.

\subsection{Results of the Model ensemble} 
\begin{table}[t]
\caption{Performance (acc. accuracy, pr. precision, rc. recall) on the West Nile Disease case study, for different methods and pre-training strategies.}
\begin{center}
        \begin{tabular}{lccccc}
        \toprule
         \mytabheader{Input} & \mytabheader{pre-training} & \mytabheader{acc.} & \mytabheader{pr.} & \mytabheader{rc.} & \mytabheader{F1} 
         \vspace{0.1cm}\\
         \hline\addlinespace[0.05cm]
         Random & \multirow{2}{*}{-} & \multirow{2}{*}{.503} & \multirow{2}{*}{.391} & \multirow{2}{*}{.395} & \multirow{2}{*}{.393} \\
         classifier & & & & & \\
         \midrule
         RGB & from scratch & .652 & .542 & .941 & .688 \\
         RGB & ImageNet & .865 & .819 & .857 & .838 \\
         \midrule
        $B_{1,8A,11,12}$ & from scratch & .756 & .662 & .817 & .732 \\
        $B_{1,8A,11,12}$ & Colorization & .852 & .823 & .811 & .817 \\
         \midrule
         Ensemble & Color.+ImageNet & \textbf{.880} & \textbf{.855} & \textbf{.850} & \textbf{.852}\\
         \bottomrule
        \end{tabular}
    \label{tab:esaresult}
    \end{center}
\end{table}
\begin{table}[t]
\caption{Comparison between several baselines and our ensemble method on \bigearth{}.}
    \begin{center} 
        \begin{tabular}{lccccc}
        \toprule
         \mytabheader{Method} & \mytabheader{pr.} & \mytabheader{rc.} & \mytabheader{F1} \vspace{0.1cm}\\
         \midrule
         K-Branch CNN~\cite{sumbul2019bigearthnet} & .716 & \textbf{.789} & .727 \\
         VGG19~\cite{sumbul2019bigearthnet} & .798 & .767 & .759 \\
         \resnet{50}~\cite{sumbul2019bigearthnet} & .813 & .774 & .771 \\
         \resnet{101}~\cite{sumbul2019bigearthnet} & .801 & .774 & .764 \\
         \resnet{152}~\cite{sumbul2019bigearthnet} & .817 & .762 & .765 \\
         \midrule
         Ensemble (our) & \textbf{84.30} & 78.10 & \textbf{81.10} & \\
         \bottomrule
        \end{tabular}
    \label{tab:sota}
    \end{center}
\end{table}

Here, we primarily assess the effectiveness of the ensemble discussed in Sec.~\ref{sec:ensemble} on \bigearth{}. In this regard, Tab.~\ref{tab:ensembleresult} compares the performance that can be reached when leveraging a twofold source of information (RGB and spectral bands): firstly, the ensemble model largely outperforms those that consider a single input modality; secondly, colorization presents an improvement over the ImageNet pre-training. 

Tab.~\ref{tab:esaresult} reports the results achieved on the West Nile Disease case study discussed in Sec~\ref{sec:westnile}. To provide a better understanding, we additionally furnish a simple baseline (\textit{i.e.} \quotationmarks{random classifier}) that computes predictions by randomly guessing from the class-prior distribution of the training set. As a first remark, all the networks we trained exceed random guessing, hence suggesting they effectively learned meaningful and suitable features for the problem at hand. Secondly, the ensemble model plays an important role even in this case, surpassing networks based on a single modality by a consistent margin.

\subsection{Comparison with the state of the art} 

To further highlight the contributions of our proposal, we compare it with the networks discussed in~\cite{sumbul2019bigearthnet}. Results reported in Tab.~\ref{tab:sota} confirm the above intuitions: the ensemble we build upon \resnet{18} outperforms heavier and overparametrized networks like \resnet{101} or \resnet{152}. Notably, we found a large improvement in precision, suggesting that our proposal is capable of returning only the categories that are relevant to the semantics of the input tile. 

It is noted the fairness of the comparisons above, as both our ensemble and the baselines leverage the same amount of information in input (namely, spectral bands and color channels). Nevertheless, an important difference subsists in the way information is consumed: while~\cite{sumbul2019bigearthnet} stacks both the input modalities to form a single input tensor, we distinguish two independent paths that eventually cross in the output space. This way, we can benefit from two different pre-training, each one being devoted to its modality: the one offered by colorization -- which works well for spectral bands -- and the ImageNet one -- which instead represents a natural and reasonable choice for dealing with RGB images. 

\subsection{Model Explanation - Towards diverse feature sets}

\begin{figure}[t]
\centering
\setlength{\tabcolsep}{1pt}
\begin{tabular}{p{0.02\textwidth}p{0.02\textwidth}cccc} 
& & \multirow{2}{*}{Urban fabric} & \multirow{2}{*}{Arable land} & Marine & Coniferous \\
& & & & waters & forest\\
\noalign{\vskip 1mm}
\rotatebox{90}{\ \textbf{Real}} & \rotatebox{90}{\ \textbf{RGB}} & \tinyrgbGT{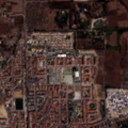} &  \tinyrgbGT{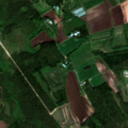} & \tinyrgbGT{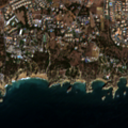} & \tinyrgbGT{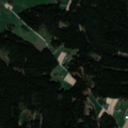} \\\hline
\noalign{\vskip 1mm}
\rotatebox{90}{\ \textbf{Predicted}} & \rotatebox{90}{\ \textbf{RGB}} & \tinypred{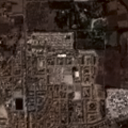} &  \tinypred{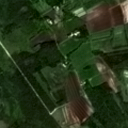} & \tinypred{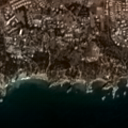} & \tinypred{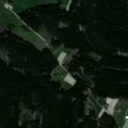} \\\hline
\noalign{\vskip 1mm}
\rotatebox{90}{\ \textbf{Expl. RGB}} & \rotatebox{90}{\ \textbf{branch}} & \tinyrgbheat{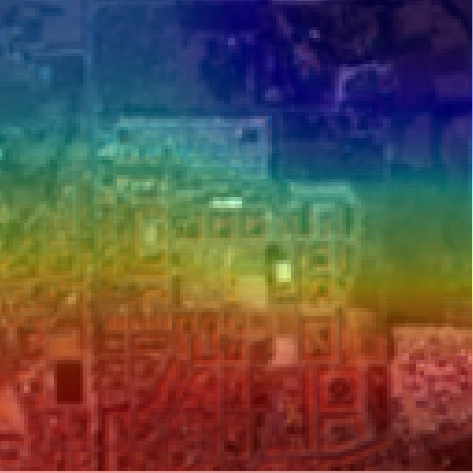} &  \tinyrgbheat{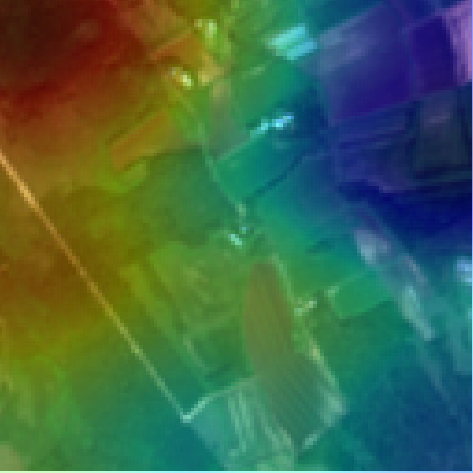} & \tinyrgbheat{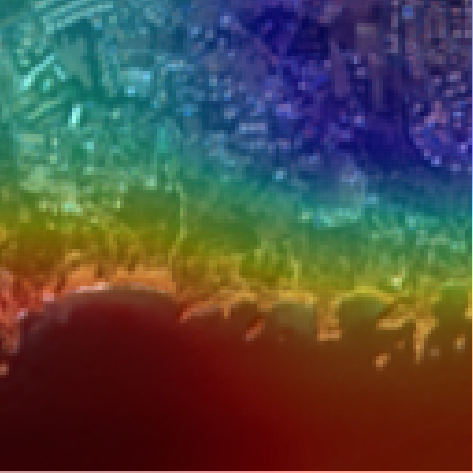} & \tinyrgbheat{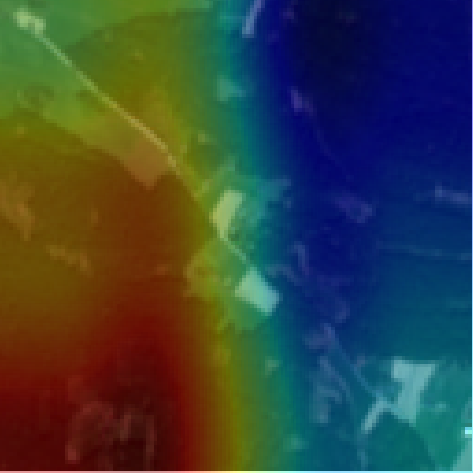} \\\hline
\noalign{\vskip 1mm}
\rotatebox{90}{\ \textbf{Heat Spec.}} &  \rotatebox{90}{\ \textbf{branch}} & \tinyspectralheat{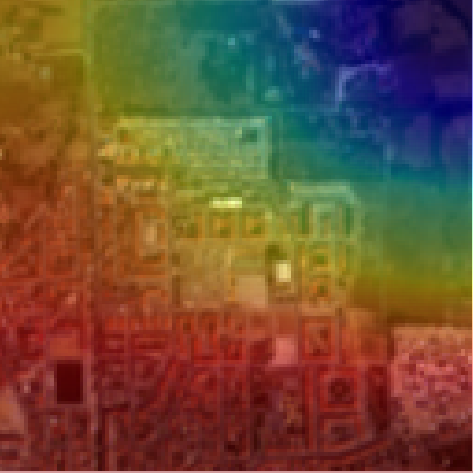} & \tinyspectralheat{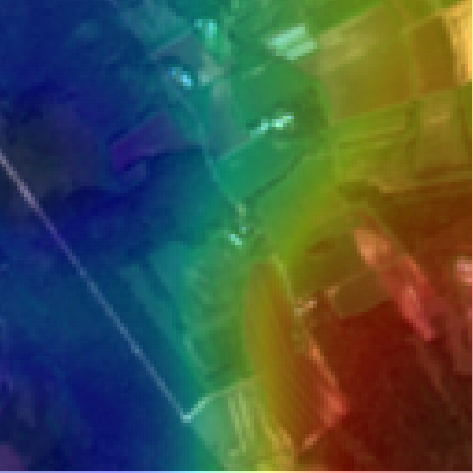} & \tinyspectralheat{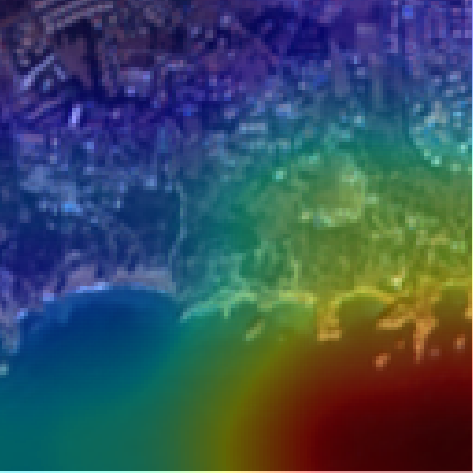} & \tinyspectralheat{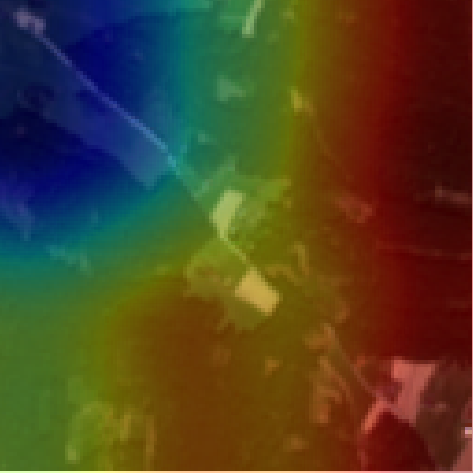} \\\hline
\end{tabular}
\caption{Some examples of the \bigearth{} dataset, coupled with the predicted colorization and visual explanations provided by the ensemble method for RGB and spectral inputs.}
\label{fig:colorexamples}
\end{figure}

We believe the strength of our ensemble approach being a result of the diversity among the individual learners. We investigate the truthfulness of such a claim from a \textit{model explanation} perspective, questioning which information in the input makes our models arrive at their decisions~\cite{samek2017explainable}. In particular, we take advantage of GradCam~\cite{selvaraju2017gradcam} to assess whether the two branches look for different properties within their inputs. The third and fourth rows of Fig.~\ref{fig:colorexamples} highlight the input regions that have been considered important for predicting the target category (we limit the analysis to the class denoting the highest confidence score). As one can see, the explanations provided by the two branches visually diverge, thus qualitatively confirming the weak correlation between their representations.

%% file: text/6_conclusions.tex
\section{Conclusion}
In this work, we propose a self-supervised learning approach for satellite imagery, which moves towards a proper initialization for deep networks facing up to Remote Sensing tasks. Our proposal builds upon two steps: firstly, we ask an encoder-decoder architecture to predict color channels from those capturing spectral information (colorization); secondly, we exploit its encoder as a pre-trained feature extractor for a classification task (\textit{i.e.} land-cover categorization and the West Nile Disease case study). We observe that the initialization we devised leads to remarkable results, exceeding the baselines especially in presence of scarce labeled data. Moreover, we qualitatively observe that representations learned through colorization are different from the ones driven by the RGB channels. Based on this finding, we set up an ensemble model that achieves the highest results in all the scenarios under consideration.